\documentclass{article}

\usepackage[dblblindworkshop, final]{neurips_2026}

\workshoptitle{Long-Context Foundation Models}

\usepackage[utf8]{inputenc} 
\usepackage[T1]{fontenc}    
\usepackage{hyperref}       
\usepackage{url}            
\usepackage{booktabs}       
\usepackage{amsfonts}       
\usepackage{nicefrac}       
\usepackage{microtype}      
\usepackage{xcolor}         
\usepackage{algorithm}
\usepackage{bbding}
\usepackage{algpseudocode}
\usepackage{amsmath}
\usepackage{threeparttable}
\usepackage{multirow}
\usepackage{hyperref}
\usepackage{graphicx}
\usepackage{amssymb} 
\usepackage{makecell}
\usepackage{subcaption}
\usepackage{tabularx}
\usepackage[compact]{titlesec}
\usepackage{enumitem}
\setlist{noitemsep, topsep=1pt, parsep=0pt, partopsep=0pt, leftmargin=*}
\usepackage[font=small,labelfont=bf,skip=5pt]{caption}
\title{LS-AR: Future-Predictive Latent Steering in Autoregressive LLMs}

\author{%
  Anubha Gupta \\
  Electronic \& Electrical Engineering\\
  University College London\\
  London, UK\\
  \texttt{anubha.dce@gmail.com} \\
  \And
  Eduardo Pignatelli \\
  Electronic \& Electrical Engineering\\
  University College London\\
  London, UK\\
  \texttt{e.pignatelli@ucl.ac.uk} \\
}

\begin{document}

\maketitle

\begin{abstract}
Standard autoregressive (AR) models process high-level task instructions, state history, and transient tokens within a single shared sequence of tokens. Consequently, they lack the architectural mechanisms needed to isolate macro-objectives from context noise. To overcome this single-channel limitation, we introduce \textbf{Latent-Steered Autoregressive (LS-AR)}, a dual-channel architecture that decouples continuous goal steering from discrete token decoding via FiLM conditioning. We evaluate a \textit{Static Goal Encoder} ($P_0$) for persistent macro-objective retention across long rollouts and a \textit{Dynamic State Tracker} ($P_t$) for recurrent latent updates during generation. On long-horizon retrieval past context limits ($H{=}1024, W{=}500$), LS-AR (Static) achieves $100\%$ target recall where parameter-matched baselines collapse ($0\%$), while increasing throughput by $\sim 35\%$ and cutting peak VRAM by $52.8\%$. In Blocksworld planning under forced perturbations ($k{=}1$), LS-AR (Dynamic) sustains an $89.0\%$ completion rate vs. $71.0\%$ for the baseline, though zero-shot entity scaling ($N \to N+1$) exposes single-vector capacity limits ($0\%$). Finally, dual-channel authority analysis shows that text goal dropout establishes latent-dominant control, offering structural defence against text prompt injection while introducing a latent vector attack surface. Code is available at \href{https://github.com/agupta1024/future-conditional-llms}{github.com/agupta1024/future-conditional-llms}.

\end{abstract}

\section{Introduction}
While autoregressive language models excel at open-ended sequence completion, they face a fundamental structural mismatch when applied to goal-directed planning. Because these models condition decoding exclusively on context window, they are forced to process macro-objectives within the exact same token stream used for local, step-by-step token generation. As sequence lengths expand during multi-step execution, this single-channel paradigm suffers from severe attention dilution and context eviction \citep{liu2024lost, xiao2024efficient}: overarching task instructions are either pushed beyond the active context window or diluted across long attention spans. Consequently, models gradually lose sight of primary goals, prioritising immediate local predicates at the expense of overall plan coherence \citep{87aa52e847784bbe8ff736aa7827d285}. Existing mitigations, such as context window scaling \citep{chen2023extending, child2019generating} or external memory architectures \citep{chhikara2025mem0, ho2025arcmemo}, incur substantial computational overhead and retrieval latency without resolving the underlying single-channel constraint.

To overcome this structural bottleneck, we introduce the \textit{Latent-Steered Autoregressive} (LS-AR) architecture, a framework that introduces a dedicated continuous channel to causal decoders to evaluate macro-objective retention over extended generation horizons. Inspired by Joint Embedding Predictive Architectures (JEPA) \citep{dawid2024introduction}, we adopt a non-generative, latent-predictive objective to model macro-intent. Rather than predicting future trajectory steps in token space, LS-AR trains a goal predictor to forecast target state representations directly in latent space ($\mathbb{R}^d$). By injecting these latent vectors into the decoder's residual stream via Feature-wise Linear Modulation (FiLM) layers \citep{perez2018film}, we architecturally decouple continuous macro-objective steering from discrete token decoding. We evaluate two operational paradigms: a \textit{Static Goal Encoder} ($P_0$) designed to bypass context eviction across extended sequence horizons, and a \textit{Dynamic State Tracker} ($P_t$) that recurrently updates the latent steering vector to absorb intermediate trajectory perturbations.

\textbf{Experimental scope.} To isolate the architectural impacts of latent steering from confounding variables such as Byte-Pair Encoding (BPE) fragmentation or opaque pre-training priors our empirical evaluation is strictly bounded to synthetic, domain-tokenised environments (procedural \textit{Treasure Hunt} and interactive \textit{Blocksworld}). We benchmark LS-AR against parameter-matched autoregressive baselines across controlled 32M and 85M parameter scales.

\textbf{Summary of contributions.}
\begin{itemize}[leftmargin=*]
    \item \textbf{Latent-Steered Autoregressive (LS-AR) framework:} We present a JEPA-inspired architecture that routes latent vectors into autoregressive decoders via FiLM layers, establishing a fixed-dimensional alternative to temporal context scaling.
    \item \textbf{Context expiration mitigation ($P_0$):} On procedural retrieval tasks, the \textit{Static} variant maintains $100\%$ target recall over extended horizons ($H = 1024$, context window $W = 500$) where parameter-matched standard AR baselines collapse to $0\%$, while reducing peak VRAM by up to $52.8\%$ and increasing throughput by $\sim 35\%$.
    \item \textbf{Dynamic resilience and error recovery ($P_t$):} In interactive Blocksworld environments, the \textit{Dynamic} variant continuously updates latent predictions to absorb trajectory perturbations, achieving an $89.0\%$ task completion rate compared to $71.0\%$ for the baseline under forced single-error recovery ($k = 1$).
    \item \textbf{Capacity limits of latent steering:} We demonstrate that scaling out-of-distribution entity counts ($N \to N+1$) causes complete task failure ($0.0\%$ success), exposing the fundamental capacity bounds of single-vector steering.
    \item \textbf{Security implications of channel dominance:} Through goal-swapping and blank-goal ablations, we analyse the control hierarchy between textual and latent channels. We reveal a new attack surface: training curricula featuring textual goal dropout grant dominant authority to the steering channel, allowing latent vectors to covertly override explicit text instructions.
\end{itemize}

\vspace{-5pt}
\section{Related work}
\vspace{-5pt}
\subsection{Context extensions and recurrent architectures}
\vspace{-5pt}
Long-horizon degradation in autoregressive models is typically framed as a context capacity constraint \citep{mohtashami2023landmark}. Mitigation strategies rely on scaling causal attention windows \citep{chen2023extending, child2019generating, liu2024ringattention} or augmenting decoders with external memory and KV-cache eviction policies \citep{lewis2020retrieval, xiao2024efficient}. However, these approaches incur quadratic $O(N^2)$ compute costs or rely on precise local retrieval queries without providing structural goal isolation. Selective State Space Models (SSMs) like Mamba \citep{gu2021efficiently, gu2023mamba} avoid quadratic costs but suffer from signal dilution, mixing macro-instructions and microscopic syntax within a single recurrent state. In contrast, LS-AR decouples macro-objectives into a dedicated steering channel, preventing target intent from being overwritten during local token generation.

\vspace{-5pt}
\subsection{Explicit planning and agentic reasoning}
\vspace{-5pt}
Autoregressive planning on multi-step benchmarks \citep{valmeekam2023planbench} often fails due to causal myopia, repeatedly trapping models in local optima such as the Sussman Anomaly \citep{sussman1973computational}. Reasoning frameworks like Chain-of-Thought \citep{wei2022chain}, Tree-of-Thoughts \citep{yao2023tree}, and ReAct \citep{yao2022react} force models to verbalise plans in token space. Over extended rollouts, these text scratchpads consume context window capacity and remain susceptible to context eviction and task amnesia. LS-AR removes macro-goal steering from the token stream entirely: target states are projected into continuous latent vectors, freeing textual generation to focus strictly on local step execution while preserving global goal alignment.

\vspace{-5pt}
\subsection{Predictive architectures (JEPA) and latent conditioning}
\vspace{-5pt}
Our methodology builds on Joint Embedding Predictive Architectures (JEPA) \citep{dawid2024introduction, assran2023self, assran2025v}, which predict abstract representations rather than exact future tokens. While LLM-JEPA \citep{huang2025llm} restricts latent prediction to an auxiliary training objective, and Agentic-JEPA \citep{rodrigogines2026agenticjepa} isolates latent representations for offline state simulation, LS-AR actively integrates these predictions into the decoding loop. By injecting the predictions directly into the decoder's residual stream via FiLM layers \citep{perez2018film}, LS-AR converts latent representations into dynamic steering vectors for autoregressive generation.

\vspace{-5pt}
\section{Methodology}
\vspace{-5pt}
\subsection{Problem formulation: asymmetric modalities and context limits}
\vspace{-5pt}
Goal-directed planning requires mapping two temporally asymmetric modalities into a joint latent space ($\mathbb{R}^d$): (1) a discrete prompt $g \in \mathbb{V}^k$ specifying initial constraints and target states (\texttt{[GOAL]}), and (2) a discrete execution trajectory $x_{\text{future}} \in \mathbb{V}^m$ satisfying $g$. Standard autoregressive models factorise trajectory generation over a sliding context window of size $W$:
\begin{equation}
P(x) = \prod_{i=1}^{|x|} P\left(x_i \;\middle|\; x_{\max(1, i-W):i-1}\right)
\end{equation}
When rollout length exceeds $W$, prompt tokens $g$ fall outside active causal attention ($i - W > k$), inducing context eviction and plan drift. LS-AR bypasses window constraints by maintaining a steering vector $P \in \mathbb{R}^d$ that persists across the decoding horizon without consuming context slots.

\vspace{-5pt}
\subsection{Static goal encoding ($P_0$)}
\vspace{-5pt}
\label{sec:vicreg}

\begin{figure}[htbp]
\centering
\includegraphics[width=0.82\linewidth]{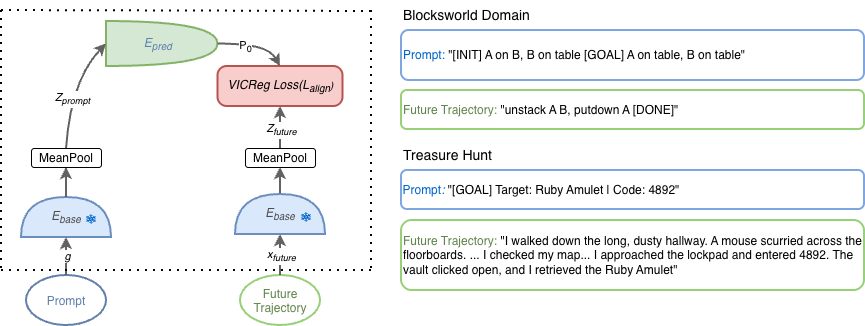}
\caption{\textbf{Static predictor alignment (Phase 2).} Predictor $E_{\text{pred}}$ maps mean-pooled prompt representations to target trajectory embeddings $Z_{\text{future}} = \text{MeanPool}(E_{\text{base}}(x_{\text{future}}))$ via VICReg loss while keeping base encoder $E_{\text{base}}$ frozen.}
\label{fig:predictor}
\end{figure}

Given a discrete prompt $g$, a frozen base encoder $E_{\text{base}}$ extracts sequence representations that are mean-pooled into an order-invariant embedding. Predictor network $E_{\text{pred}}$ projects this embedding into the latent space $\mathbb{R}^d$:
\begin{equation}
P_0 = E_{\text{pred}}\Big(\text{MeanPool}\big(E_{\text{base}}(g)\big)\Big)
\end{equation}
During Phase 2 alignment training (Figure~\ref{fig:predictor}), $E_{\text{pred}}$ is optimised to match the target representation of the ground-truth execution rollout:
\begin{equation}
Z_{\text{future}} = \text{MeanPool}\big(E_{\text{base}}(x_{\text{future}})\big) \in \mathbb{R}^d
\end{equation}

\vspace{-5pt}
\subsection{Dynamic state tracker ($P_t$)}
\vspace{-5pt}

\begin{figure}[htbp]
\centering
\includegraphics[width=0.82\linewidth]{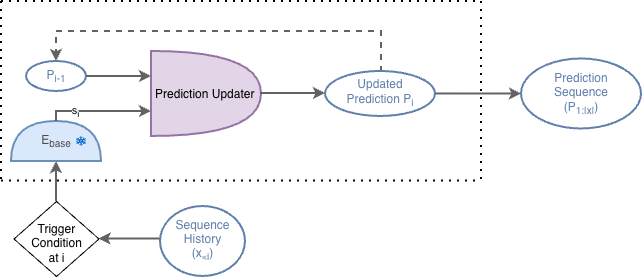}
\caption{\textbf{Dynamic state tracker.} At trigger steps $i$, the tracker updates the latent $P_{i-1}$ using hidden state $s_i = E_{\text{base}}(x_{1:i})_i$ to form step-aligned prior $P_i$.}
\label{fig:updater}
\end{figure}
To preserve alignment under environmental perturbations, the Dynamic State Tracker (Figure~\ref{fig:updater}) updates the latent steering vector at discrete state transitions. See Appendix~\ref{sec:pool} for the architectural rationale behind trigger boundary and pooling choices. When token $x_i$ matches a pre-defined trigger token, execution history $x_{1:i}$ is processed through frozen $E_{\text{base}}$ to extract trigger-position state $s_i = E_{\text{base}}(x_{1:i})_i \in \mathbb{R}^d$.

A Multi-Head Attention (MHA) block computes state context $A_i = \text{MHA}(Q = P_{i-1}, K = s_i, V = s_i)$. Although softmax over a single key collapses to $1.0$, retaining MHA preserves multi-subspace projections and supports multi-token state histories. $A_{i}$ is passed through a Multi-Layer Perceptron (MLP) scaled by a learnable scalar $\alpha \in \mathbb{R}$. The active prediction vector $P_{i}$ updates conditionally at step $i$:
\begin{equation}
P_i = \begin{cases} 
P_{i-1} + \alpha \cdot \Big[ A_i + \text{MLP}\big(\text{LayerNorm}(P_{i-1} + A_i)\big) \Big], & \text{if } x_i = \text{Trigger Token} \\ 
P_{i-1}, & \text{otherwise} 
\end{cases}
\end{equation}
Initialised at $\alpha = 0$ \citep{bachlechner2021rezero}, the tracker acts as an identity function at step zero, learning corrective updates only when observed trajectory states diverge from expectations. Across all sub-modules ($E_{\text{pred}}$, $E_{\text{base}}$, FiLM, and MHA), we fix $d = 512$ to avoid intermediate projection layers. Consequently, while standard token generation retains $O(1)$ complexity, state tracking incurs a sparse $O(i)$ encoding step exclusively at trigger boundaries.

\vspace{-5pt}
\subsection{Latent steering via FiLM}
\vspace{-5pt}
\begin{figure}[htbp]
\centering
\includegraphics[width=0.82\linewidth]{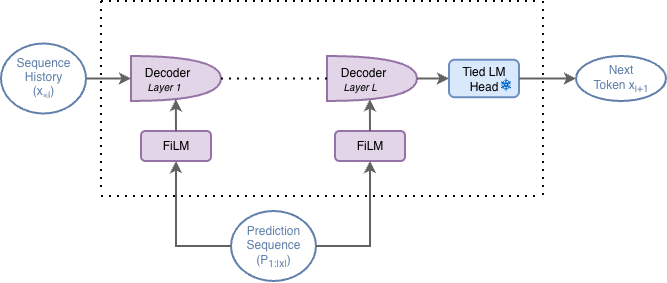}
\caption{\textbf{Conditioned decoding.} At step $i$, steering vector $P_i$ modulates decoder layer normalisations via FiLM scale $\gamma$ and shift $\beta$ parameters to condition token prediction $x_{i+1}$.}
\label{fig:decoder}
\end{figure}
LS-AR injects the steering vector $P_i$ ($P_0$ for Static; $P_{1:|x|}$ for Dynamic) directly into the causal decoder, $D_{\text{base}}$, using layer-wise Feature-wise Linear Modulation (FiLM) (Figure~\ref{fig:decoder}). For decoder layer $l$ and position $i$, affine parameters are linearly projected from $P_i$:
\begin{equation}
\gamma_{l,i} = W_{\gamma,l} P_i + b_{\gamma,l} \quad \text{and} \quad \beta_{l,i} = W_{\beta,l} P_i + b_{\beta,l}
\end{equation}
FiLM modulates the pre-MLP normalisation of post-attention hidden state $h_{l,i}$:
\begin{equation}
\tilde{h}_{l,i} = \text{LayerNorm}_2(h_{l,i}) \odot (1 + \gamma_{l,i}) + \beta_{l,i}
\end{equation}
Zero-initialisation of $W_{\gamma, \beta}$ and $b_{\gamma, \beta}$ ensures FiLM acts as an identity transform at initialisation, preserving pre-trained model stability.

\vspace{-5pt}
\subsection{Phased training curriculum}
\label{sec:training_curriculum}
\vspace{-5pt}
LS-AR is trained across three distinct phases:

\begin{itemize}[leftmargin=*]
    \item \textbf{Phase 1. Base domain adaptation:} Causal fine-tuning on domain syntax to establish base vocabulary and transition dynamics.
    \item \textbf{Phase 2 (Predictor alignment, $\mathcal{L}_{\text{align}}$):} Frozen $E_{\text{base}}$ provides targets for $E_{\text{pred}}$, optimised via non-contrastive VICReg loss \citep{bardes2021vicreg}:
    \begin{equation}
    \mathcal{L}_{\text{align}} = \lambda_s \, s(P_0, Z_{\text{future}}) + \lambda_v \big[ v(P_0) + v(Z_{\text{future}}) \big] + \lambda_c \big[ c(P_0) + c(Z_{\text{future}}) \big]
    \end{equation}
    where $s$ denotes mean squared error, $v$ maintains variance across batch dimensions, and $c$ decorrelates features (see Appendix~\ref{sec:vicreg_hyper} for full mathematical definitions).
    \item \textbf{Phase 3 (Conditioned generation, $\mathcal{L}_{\text{AR}}$):} Decoder $D_{\text{base}}$, FiLM layers, and Dynamic State Tracker are trained end-to-end on next-token prediction:
    \begin{equation}
    \mathcal{L}_{\text{AR}} = -\sum_{i=1}^{|x|} \log P\left(x_i \;\middle|\; x_{<i}, P_{i-1}\right)
    \end{equation}
    \item \textbf{Phase 3b (Textual goal dropout):} In $30\%$ of Phase 3 steps, textual prompt $g$ is replaced with padding tokens. This forces $D_{\text{base}}$ to rely entirely on $P_i$, transferring primary steering authority to the latent channel.
\end{itemize}
Complete training configurations and hyperparameter schedules are provided in Appendix \ref{sec:train_hyper}.

\vspace{-5pt}
\section{Experiments}
\vspace{-5pt}

To evaluate latent steering independently of orthogonal prompting or retrieval mechanisms (e.g., Chain-of-Thought or RAG), we compare LS-AR directly against parameter-matched autoregressive causal decoders across strictly controlled parameter footprints, tokeniser vocabularies, and training corpora.

\vspace{-5pt}
\subsection{Mitigating context expiration (Treasure Hunt)}
\vspace{-2pt}
\paragraph{Experimental setup \& baselines} 

We benchmark \textbf{LS-AR (Static)} on fixed-target procedural retrieval using the \textit{Treasure Hunt} dataset ($50\text{k}$ training trajectories, $500$ evaluation instances; Appendix~\ref{sec:th_dataset}), which combines elements of Needle in a Haystack \citep{nelson2024needle} and Passkey retrieval \citep{mohtashami2023landmark}. Using standard BPE tokenisation ($|V| = 50,257$), the parameter-matched AR baseline employs a 20-layer causal decoder ($\sim 89\text{M}$ parameters). LS-AR (Static) matches this capacity ($\sim 85\text{M}$ parameters) using a 4-layer base encoder ($E_{\text{base}}$), a 4-layer causal decoder ($D_{\text{base}}$), weight-tied embeddings, and per-layer FiLM projection heads.

In each test instance, a prompt specifying a target item (e.g., \textit{Ruby Amulet}) and a 4-digit passkey (e.g., \textit{4892}) is followed by an irrelevant distractor sequence of length $H$. We enforce prefix forcing -- injecting distractor tokens before terminal retrieval -- and set the context window $W = 500$ across expanding horizons $H \in \{64, 128, 256, 512, 1024\}$. When $H > W$, prompt tokens slide out of active causal attention, testing recall beyond the context boundary. Performance is measured via exact match on \textbf{Semantic Target Retrieval} (macro-goal routing) and \textbf{Numeric Passkey Retrieval} (4-digit sub-word BPE accuracy).



\begin{table}[htbp]
\centering
\caption{\textbf{Semantic and numeric memory retrieval ($W = 500$)} across in-context ($H=128$) and out-of-context ($H=1024$) generation horizons. Values represent $\text{Mean} \pm \text{SEM}$ across 5 random seeds.}
\label{tab:passkey_retention}
\resizebox{\columnwidth}{!}{
\begin{tabular}{llcc}
\toprule
\textbf{Horizon ($H$)} & \textbf{Architecture} & \textbf{Target Retrieval (Semantic) $\uparrow$} & \textbf{Passkey Retrieval (Numeric) $\uparrow$} \\
\midrule
\multirow{2}{*}{\shortstack[l]{$H=128$\\ (In-Context)}} & Baseline (AR GPT-2)  & $100.0 \pm 0.0\%$ & $39.8 \pm 4.44\%$ \\
& LS-AR (Static)  & $\mathbf{100.0 \pm 0.0\%}$ & $\mathbf{92.8 \pm 1.48\%}$  \\
\midrule
\multirow{2}{*}{\shortstack[l]{$H=1024$\\ (Out-of-Context)}} & Baseline (AR GPT-2)  & $0.0 \pm 0.0\%$ & $0.0 \pm 0.0\%$  \\
& LS-AR (Static)  & $\mathbf{100.0 \pm 0.0\%}$ & $\mathbf{84.4 \pm 4.51\%}$ \\
\bottomrule
\end{tabular}
}
\end{table}

\begin{figure}[htbp]
     \centering
     \begin{subfigure}[b]{0.45\linewidth}
         \centering
         \includegraphics[width=\linewidth]{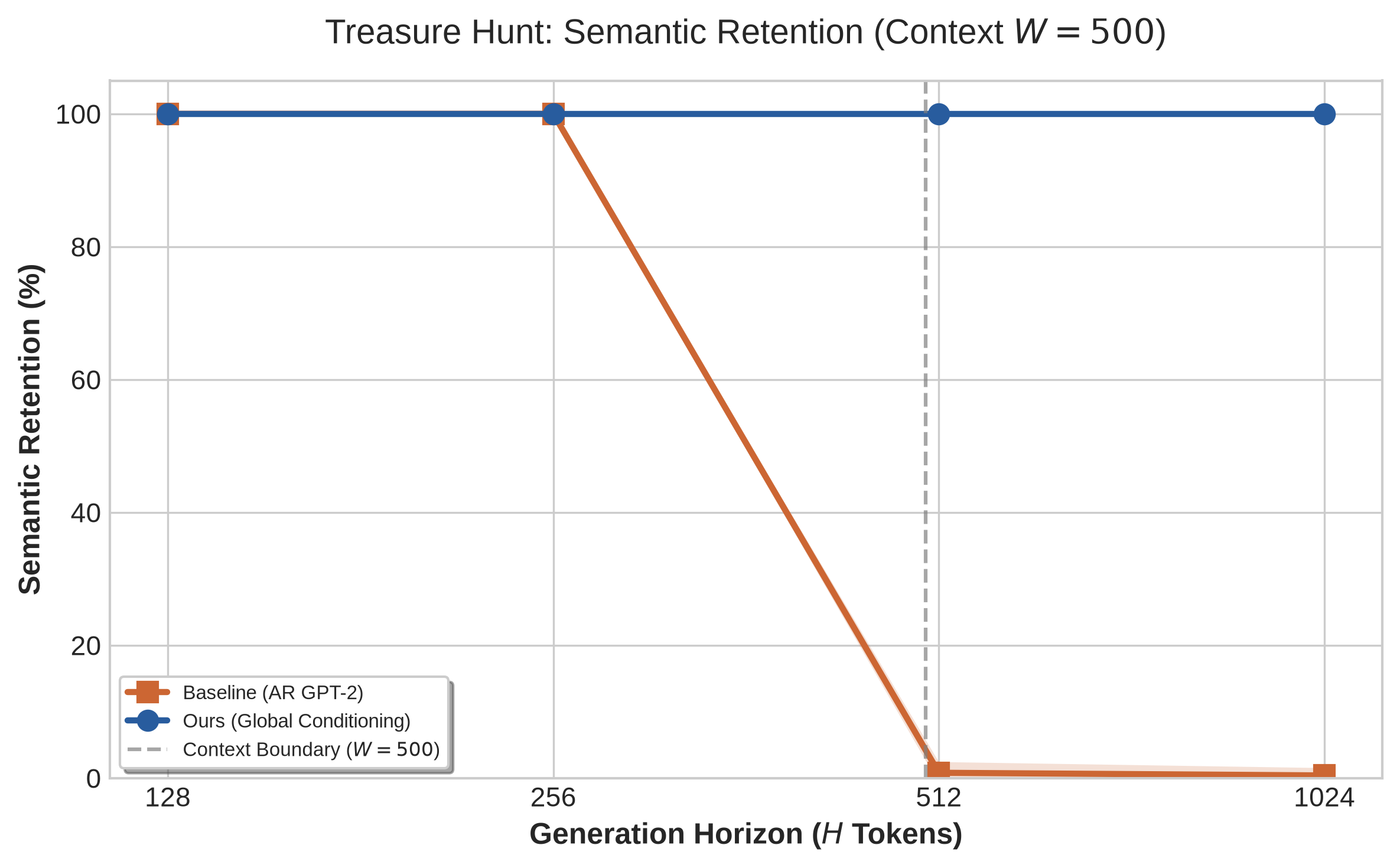}
         \caption{}\label{fig:th_target_retention}
     \end{subfigure}
     \hfill
     \begin{subfigure}[b]{0.52\linewidth}
         \centering
         \includegraphics[width=\linewidth]{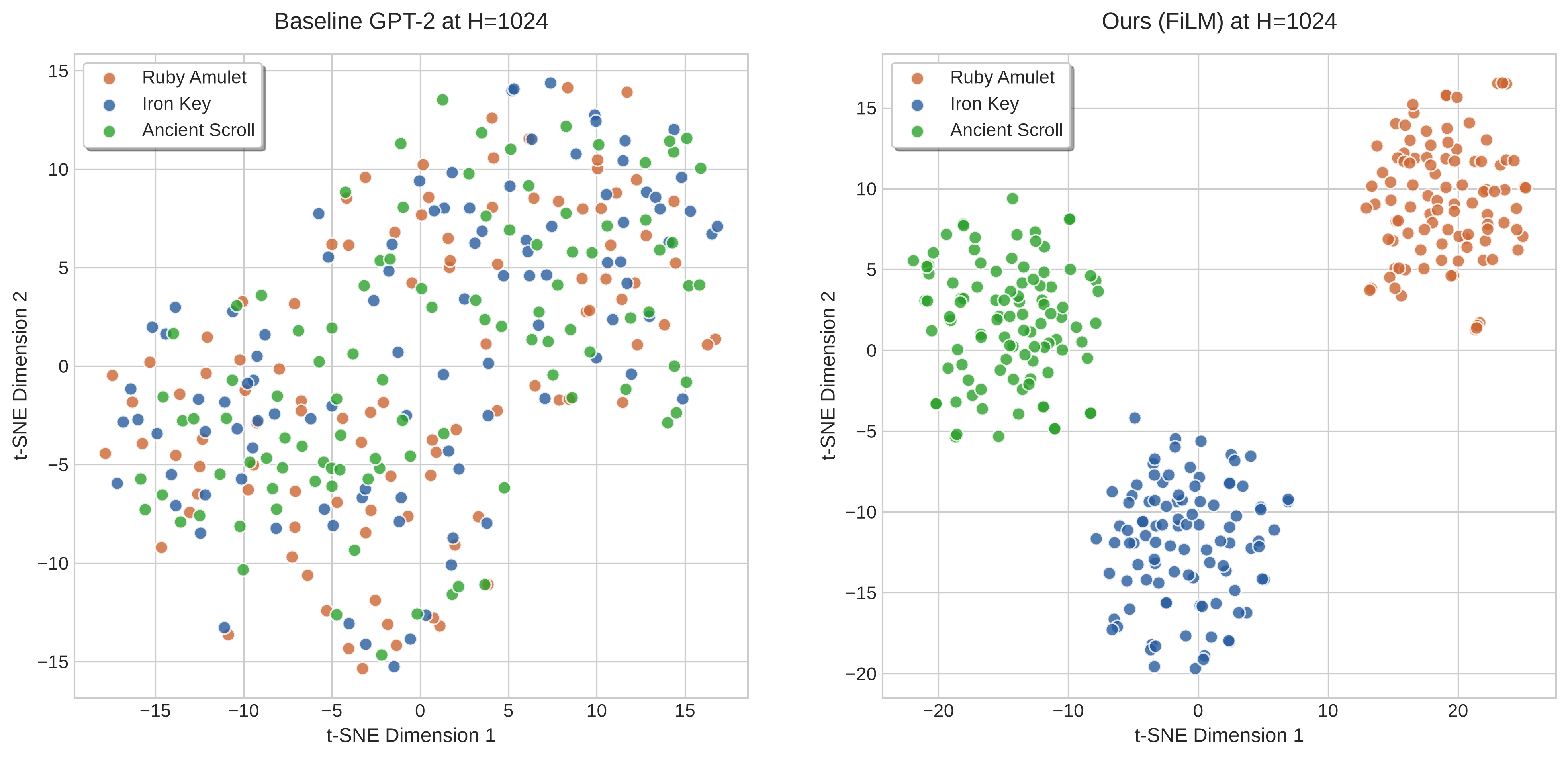}
         \caption{}\label{fig:th_tsne}
     \end{subfigure}
     \caption{\textbf{Performance and representational analysis across context boundary ($W=500$).} \textbf{(a)} Semantic target retrieval across expanding horizons ($H$). The baseline collapses when $H > W$, while LS-AR maintains $100\%$ retrieval regardless of sequence length. \textbf{(b)} t-SNE of decoder hidden states immediately preceding terminal retrieval at $H=1024$ ($N=100$). Baseline states collapse into distractor noise, whereas LS-AR forms tightly separated clusters by target identity.}
     \label{fig:ts_snr}
\end{figure}

\paragraph{Results and representational analysis}
\vspace{-5pt}
As shown in Table~\ref{tab:passkey_retention} and Figure~\ref{fig:th_target_retention}, the AR baseline achieves $100.0\%$ semantic target retrieval within context ($H=128$), but struggles with exact numeric passkey extraction ($39.8\%$) due to sub-word token fragmentation across distractor text. At $H=1024$, it suffers complete amnesia ($0.0\%$). Conversely, LS-AR (Static) maintains $100.0\%$ semantic recall across all horizons, proving that continuous steering insulates macro-objectives from context eviction. Passkey retrieval shows minor degradation at $H=1024$ ($84.4\%$), reflecting slight loss when compressing discrete digit sequences into continuous embeddings. t-SNE visualisation (Figure~\ref{fig:th_tsne}) confirms that baseline hidden states collapse into local distractor noise at $H=1024$, whereas LS-AR preserves tightly organised clusters by target identity.

\vspace{-5pt}
\subsection{Computational efficiency of static conditioning}
\vspace{-5pt}
We benchmark autoregressive KV-cached decoding efficiency across expanding context limits ($W \in \{128, 256, 500\}$) against two bounds: a parameter-matched \textbf{Baseline-Large} (20-layer decoder, $89\text{M}$) and a decoder-matched \textbf{Baseline-Small} (4-layer unconditioned decoder, $38\text{M}$).

\begin{table}[h]
\centering
\caption{\textbf{Decoding efficiency of LS-AR (Static)} across context windows ($W$). $\Delta$ values are anchored strictly against the parameter-matched Baseline-Large. Values represent $\text{Mean} \pm \text{SEM}$ across 5 random seeds.}
\label{tab:efficiency_static}
\resizebox{\columnwidth}{!}{%
\begin{tabular}{llcccc}
\toprule
\textbf{Context Size} & \textbf{Architecture} & \textbf{Throughput ($\uparrow$)} & \textbf{$\Delta$ Speed} & \textbf{Peak VRAM ($\downarrow$)} & \textbf{$\Delta$ Memory} \\ 
\midrule

\multirow{3}{*}{$W=128$} 
 & Baseline-Small$^*$ (38M) & $6030.46 \pm 316.21$ & -- & $1056.00 \pm 0.00$ & -- \\
 & Baseline-Large (89M Anchor) & $2381.51 \pm 56.59$ & \textit{Anchor} & $1538.74 \pm 0.00$ & \textit{Anchor} \\
 & \textbf{LS-AR (Static 85M)} & $\mathbf{3196.66 \pm 164.55}$ & \textbf{+34.23\%} & $\mathbf{1039.28 \pm 0.00}$ & \textbf{-32.46\%} \\ 
\midrule

\multirow{3}{*}{$W=256$} 
 & Baseline-Small$^*$ (38M) & $5934.56 \pm 306.78$ & -- & $1103.05 \pm 0.00$ & -- \\
 & Baseline-Large (89M Anchor) & $2190.98 \pm 35.06$ & \textit{Anchor} & $1858.41 \pm 0.00$ & \textit{Anchor} \\
 & \textbf{LS-AR (Static 85M)} & $\mathbf{3085.42 \pm 125.18}$ & \textbf{+40.82\%} & $\mathbf{1059.43 \pm 0.00}$ & \textbf{-42.99\%} \\ 
\midrule

\multirow{3}{*}{$W=500$} 
 & Baseline-Small$^*$ (38M) & $5824.46 \pm 271.55$ & -- & $1202.81 \pm 0.00$ & -- \\
 & Baseline-Large (89M Anchor) & $2075.58 \pm 26.61$ & \textit{Anchor} & $2389.06 \pm 0.00$ & \textit{Anchor} \\
 & \textbf{LS-AR (Static 85M)} & $\mathbf{2821.27 \pm 99.79}$ & \textbf{+35.93\%} & $\mathbf{1127.38 \pm 0.00}$ & \textbf{-52.81\%} \\ 
 
\bottomrule

\multicolumn{6}{l}{\small $^*$Baseline-Small represents the unconditioned 4-layer decoding throughput upper bound.}
\end{tabular}%
}
\end{table}
\paragraph{Efficiency analysis} As detailed in Table~\ref{tab:efficiency_static}, LS-AR yields a $+34.23\%$ to $+40.82\%$ throughput gain over Baseline-Large. Because Baseline-Large evaluates all 20 decoder layers at every token step, compute overhead accumulates rapidly ($2381.51 \to 2075.58\,\text{tok/s}$). By contrast, LS-AR offloads prompt extraction to a single run-once pass through $E_{\text{base}}$ and decodes via a 4-layer backbone, reducing peak VRAM by up to $-52.81\%$ ($1127.38\,\text{MB}$ vs. $2389.06\,\text{MB}$ at $W=500$).

\vspace{-5pt}
\subsection{Blocksworld: logical planning and dynamic resilience}
\vspace{-5pt}
\label{sec:bw_planning_resilience}
\paragraph{Experimental setup \& baselines}
Agentic planning demands deterministic execution, resistance to search traps (e.g., Sussman Anomaly), and error recovery under environmental perturbations. To evaluate these properties we benchmark \textbf{LS-AR (Dynamic)} on interactive logical planning using the Blocksworld domain \citep{valmeekam2023planbench} ($100\text{k}$ training trajectories, $500$ evaluation instances; Appendix~\ref{sec:bw_dataset}). To remain unburdened by large vocabulary embeddings, this experiment uses domain-specific tokenisation ($|V| = 100$; Appendix~\ref{sec:bw_tokenization}). The parameter-matched AR baseline employs a 10-layer decoder ($\sim 32\text{M}$ parameters), while LS-AR (Dynamic) matches this capacity ($\sim 32\text{M}$ parameters) using a 4-layer encoder and a 4-layer decoder.

Models are fine-tuned on trajectories with $N = 5$ blocks, where training data includes single forced sub-optimal actions ($K = 1$) followed by trajectory corrections. Evaluation spans three regimes: (1) \textbf{Optimal execution ($N=5, K=0$)}, (2) \textbf{Dynamic resilience ($K \in \{1, 2\}$)} via injected sub-optimal steps, and (3) \textbf{OOD generalisation ($N=6$)} on unseen block counts. Performance is evaluated via binary \textbf{Goal success}.

\begin{table}[htbp]
\centering
\caption{\textbf{Blocksworld planning, resilience, and OOD generalisation.} Goal success rates across optimal trajectories ($K=0$), forced perturbation steps ($K \in \{1, 2\}$), and zero-shot entity scaling ($N=6$).  Values represent $\text{Mean} \pm \text{SEM}$ across 5 random seeds.}
\label{tab:blocksworld_combined}
\resizebox{\columnwidth}{!}{%
\begin{tabular}{l|ccc|c}
\toprule
\multirow{2}{*}{\textbf{Architecture}} & \multicolumn{3}{c|}{\textbf{In-Distribution ($N=5$)}} & \textbf{OOD generalisation} \\
 & \textbf{$K=0$ (Optimal)} & \textbf{$K=1$ (1 Mistake)} & \textbf{$K=2$ (Zero-Shot)} & \textbf{$N=6$ (Zero-Shot)} \\
\midrule
Baseline (AR GPT-2) & $96.0\% \pm 0.0$ & $71.0\% \pm 0.0$ & $6.0\% \pm 0.0$ & $0.0\% \pm 0.0$ \\
\midrule
LS-AR (Static) & $95.0\% \pm 0.0$ & $66.0\% \pm 0.0$ & $5.0\% \pm 0.0$ & $0.0\% \pm 0.0$ \\
LS-AR (Dynamic) & $\mathbf{98.0\% \pm 0.0}$ & $\mathbf{89.0\% \pm 0.0}$ & $\mathbf{7.0\% \pm 0.0}$ & $0.0\% \pm 0.0$ \\
\midrule
\textit{Random FiLM Ablation} & \textit{0.0\%} & \textit{--} & \textit{--} & \textit{--} \\
\bottomrule
\end{tabular}%
}
\end{table}
\vspace{-5pt}
\paragraph{Results \& analysis} Under unperturbed execution ($K=0$), all models achieve high accuracy ($95.0\%$--$98.0\%$). Substituting the steering vectors with Gaussian noise (\textit{Random FiLM}) reduces performance to $0.0\%$, confirming that generation is strictly driven by latent steering. Under off-trajectory perturbations ($K=1$), LS-AR (Dynamic) achieves $89.0\%$ recovery, significantly outperforming the AR baseline ($71.0\%$) and LS-AR Static ($66.0\%$). Recalculating latent representations at action boundaries enables the Dynamic Tracker to absorb environmental perturbations. Multi-step errors ($K=2$) induce compounding vector drift ($5.0\%$--$7.0\%$ Goal success), and zero-shot scaling ($N=6$) yields $0.0\%$ success, showing that single-vector latent steering does not grant zero-shot extrapolation to unseen object spaces.


\vspace{-5pt}
\subsection{Dynamic state-tracking overhead}
\vspace{-5pt}

We profile the per-step tracking overhead of LS-AR (Dynamic) on a $32\text{M}$ parameter model at $W=32$, where metric overhead remains representative across context scales ($W \in \{8, 16, 32, 64\}$).

\begin{table}[htbp]
\centering
\caption{\textbf{Inference efficiency of LS-AR (Dynamic)} ($W=32$, $32\text{M}$ parameters). $\Delta$ metrics are anchored strictly against the parameter-matched Baseline-Large ($32\text{M}$). Sparsely executing $E_{\text{pred}}$ yields $+33.56\%$ higher throughput while preserving peak memory footprint. Values represent $\text{Mean} \pm \text{SEM}$ across 5 random seeds.}
\label{tab:efficiency_32m}
\resizebox{\linewidth}{!}{
\begin{tabular}{llccccc}
\toprule
\textbf{Context Size} & \textbf{Architecture} & \textbf{Decoder Depth} & \textbf{Throughput ($\uparrow$)} & \textbf{$\Delta$ Speed} & \textbf{Peak VRAM ($\downarrow$)} & \textbf{$\Delta$ Memory} \\ 
\midrule
\multirow{6}{*}{$W=32$} 
& Baseline-Small (13M) & 4 Layers & $277.5 \pm 31.7$ & -- & $431.6 \pm 0.0$ & -- \\
 & \textit{(Decoder-Matched)} & & & & & \\
\cmidrule{2-7}
 & Baseline-Large (32M) & 10 Layers & $180.0 \pm 2.7$ & \textit{Anchor} & $432.1 \pm 0.0$ & \textit{Anchor} \\
 & \textit{(Parameter-Matched)} & & & & & \\
\cmidrule{2-7}
 & \textbf{LS-AR (Dynamic 32M)} & 4 Layers (Dec) & $\mathbf{240.0 \pm 23.0}$ & \textbf{+33.56\%} & $\mathbf{433.8 \pm 0.0}$ & \textbf{0.41\%} \\
 & & \textit{+ 4 Layers ($E_{\text{base}}$)} & & & & \\
\bottomrule
\end{tabular}
}
\end{table}
\paragraph{Efficiency analysis} As shown in Table~\ref{tab:efficiency_32m}, LS-AR (Dynamic) achieves $+33.56\%$ higher throughput than the $32\text{M}$ baseline with a negligible $+0.41\%$ VRAM increase. Standard AR models evaluate all 10 decoder layers at every step, whereas LS-AR executes the deep predictor ($E_{\text{pred}}$) only at sparse trigger boundaries, keeping per-step generation bound to a lightweight 4-layer backbone.

\vspace{-5pt}
\subsection{Dual-channel control and security trade-offs}
\label{sec:bw_latent_control}
\vspace{-5pt}
To map channel authority under potential attack vectors \citep{greshake2023not, qi2024visual}, we evaluate an adversarial override regime: the decoder receives prompt text specifying \textit{Goal A}, while FiLM layers are conditioned on a latent vector encoding \textit{Goal B} (Adversarial Swap). We evaluate two training curricula: (1) \textbf{Standard curriculum} (joint text and latent training), and (2) \textbf{Latent-forced curriculum} (prompt text randomly masked with $30\%$ probability during Phase 3).

\begin{table}[htbp]
\centering
\caption{\textbf{Latent controllability and modality authority}. Under Standard training, LS-AR prioritises textual authority; under Latent-Forced training, steering vectors act as high-authority control overrides. Values represent $\text{Mean} \pm \text{SEM}$ across 5 random seeds.}
\label{tab:latent_control}
\resizebox{\columnwidth}{!}{%
\begin{tabular}{l|l|cc|cc}
\toprule
\textbf{Condition} & \textbf{Architecture} & \multicolumn{2}{c|}{\textbf{Standard Curriculum}} & \multicolumn{2}{c}{\textbf{Latent-Forced Curriculum}} \\
\textbf{Textual Goal, Injected Latent Goal} & & \textbf{Goal A Success} & \textbf{Goal B Success} & \textbf{Goal A Success} & \textbf{Goal B Success} \\
\midrule
\multirow{2}{*}{Control$^*$ ($G_A$, $G_A$)} & Baseline & $63.6\% \pm 0.94$ & 0.0\% & -- & -- \\
 & LS-AR (Dynamic) & $60.4\% \pm 1.07$ & 0.0\% & $72.2\% \pm 0.85$ & 0.0\% \\
\midrule
\multirow{2}{*}{Control$^*$ ($G_B$, $G_B$)} & Baseline & 0.0\% & $62.2\% \pm 0.85$ & -- & -- \\
 & LS-AR (Dynamic) & 0.0\% & $61.4\% \pm 1.12$ & 0.0\% & $73.2\% \pm 0.72$ \\
\midrule
Adversarial$^\dagger$ ($G_A$, $G_B$) & LS-AR (Dynamic) & $\mathbf{56.2\% \pm 1.39}$ & $0.4\% \pm 0.36$ & $1.4\% \pm 0.54$ & $\mathbf{67.2\% \pm 1.65}$ \\
Blank Goal Text (None, $G_B$)$^\dagger$ & LS-AR (Dynamic) & 0.0\% & $0.2\% \pm 0.18$ & 0.0\% & $\mathbf{71.0\% \pm 1.16}$ \\
Blank Goal Latent ($G_A$, None)$^\dagger$ & LS-AR (Dynamic) & $30.4\% \pm 2.33$ & 0.0\% & $1.8\% \pm 0.58$ & 0.0\% \\
\bottomrule
\multicolumn{6}{l}{\small $^*$ Control conditions where the Textual Goal and the Latent Goal perfectly align.} \\
\multicolumn{6}{l}{\small $^\dagger$ Standard AR baselines lack a continuous channel; latent swaps and blank ablations do not apply.} \\
\end{tabular}%
}
\end{table}
\paragraph{Authority analysis \& security trade-offs} As detailed in Table~\ref{tab:latent_control}, under the \textbf{Standard Curriculum}, the model prioritises text instructions, executing Goal A ($56.2\%$) while suppressing injected Goal B ($0.4\%$). Under the \textbf{Latent-Forced Curriculum}, steering vectors become primary control channels, overriding prompt text to execute Goal B ($67.2\%$) while suppressing Goal A ($1.4\%$).

This reveals a key security trade-off: enforcing strong latent steering via textual dropout creates vulnerability to covert steering vector injection, whereas standard joint training anchors decoding authority to human-auditable text prompts, offering structural defence against latent manipulation.

\vspace{-5pt}
\section{Discussion}
\vspace{-5pt}

The empirical results across our benchmarks demonstrate that \textbf{LS-AR} maintains semantic objective retention over extended generation horizons by leveraging continuous global conditioning. By analysing the context expiration failures of the AR baseline and the recovery mechanics of our framework, we identify broader implications for goal-directed sequence modelling.

\vspace{-5pt}
\subsection{Latent steering vs. temporal context scaling}
\vspace{-5pt}
Standard context window expansion ($W$) remains vulnerable to attention dilution and prompt eviction. By projecting macro-objectives into a persistent latent vector $P_t$, LS-AR isolates high-level goal steering from local token generation. This decoupling maintains global steering authority past active context boundaries while delivering $\sim 35\%$ throughput gains by offloading compute to a shallow 4-layer decoder backbone.

\vspace{-5pt}
\subsection{Limits of out-of-distribution structural generalisation}
\vspace{-5pt}
Single-vector conditioning exhibits clear structural capacity bounds under out-of-distribution (OOD) shift ($N=6$ block scaling yields $0\%$ success across all variants). A single fixed-dimensional latent vector lacks the relational capacity required for zero-shot combinatorial binding over unobserved object configurations.

\vspace{-5pt}
\subsection{Modality authority and security dynamics}
\vspace{-5pt}
Adversarial evaluations reveal that channel dominance -- prioritising prompt text or latent vectors -- is an engineered artifact of the training curriculum:
\begin{itemize}[leftmargin=*]
    \item \textbf{Latent-dominant regime (goal dropout):} Textual goal dropout establishes latent vectors as the primary steering channel ($67.2\%$ injected goal execution). This provides structural defence against text-based prompt injections \citep{greshake2023not}, but creates vulnerability to gradient-based vector perturbations \citep{zou2023universal, qi2024visual}.
    \item \textbf{Text-dominant regime (standard training):} Standard joint training grounds authority in prompt text ($56.2\%$ primary goal execution), suppressing unsanctioned vector swaps ($0.4\%$) and ensuring auditability.
\end{itemize}

Rather than enforcing a fixed trust model, LS-AR allows developers to tune modality authority based on deployment constraints -- selecting latent dominance when processing untrusted text, or text dominance when auditability and latent robustness are paramount.

\vspace{-5pt}
\subsection{Limitations}
\vspace{-5pt}
LS-AR exhibits four primary operational boundaries: (1) evaluation is restricted to synthetic symbolic environments, leaving continuous language transfer for future work; (2) compressing macro-intent into a single vector causes minor sub-word BPE digit scrambling during numeric passkey extraction; (3) single-vector bottlenecking bounds combinatorial resolution on OOD tasks; and (4) comparisons against unaugmented AR models serve as a preliminary bounds check, necessitating future evaluation against memory-augmented baselines (e.g., continuous KV-caches) to isolate architectural gains.

\vspace{-5pt}
\section{Conclusion and future work}
\vspace{-5pt}
We introduced \textbf{LS-AR}, an architecture that modulates autoregressive token generation with latent goal vectors via FiLM conditioning. LS-AR (Static) achieves $100\%$ semantic goal recall well beyond context limits ($H=1024, W=500$) with a $+35\%$ speedup over parameter-matched baselines, while LS-AR (Dynamic) yields $89.0\%$ recovery under off-trajectory perturbations. 

\textbf{Future Directions.} To address single-vector compression bounds, we outline three key extensions: (1) \textbf{Multi-slot latent memory:} expanding single vectors into $k$-slot matrices ($P_t \in \mathbb{R}^{k \times d}$) to enable $N \to N+1$ combinatorial scaling; (2) \textbf{Adaptive updating:} replacing fixed step triggers with entropy-guided dynamic state updates; and (3) \textbf{Open-vocabulary scaling:} evaluating latent steering on multi-billion parameter backbones across complex agentic benchmarks (e.g., ALFWorld, WebArena).

\bibliography{main}
\bibliographystyle{plainnat}


\appendix

\section{Appendix A: Architectural rationale for dynamic state tracking}
\label{sec:pool}
\subsection{Context aggregation: trigger-position vs. explicit pooling}
To construct the historical context summary $s_i$ from generated sequence $x_{1:i}$, we extract the hidden state at the trigger boundary:
$$s_i = E_{\text{base}}(x_{1:i})_i \in \mathbb{R}^d$$
We opted for this single-position summary over explicit mean- or max-pooling across the sub-action tokens for two main reasons:
\begin{itemize}[leftmargin=*]
    \item \textbf{Implicit Causal Aggregation: }Because $E_{\text{base}}$ employs causal self-attention, the token representation at the terminal position $i$ naturally attends to and integrates contextual information from all preceding tokens in the prefix $x_{1:i}$.
    \item \textbf{Computational Efficiency: }Utilising $s_i$ directly avoids storing intermediate activation matrices across token spans for auxiliary pooling, reducing memory overhead during long-trajectory generation.
\end{itemize}

\subsection{Event-driven vs. fixed-interval triggers}
\label{sec:trigger_boundary}
Instead of invoking the Dynamic State Tracker at fixed token intervals (e.g., every $N$ tokens), updates are triggered strictly upon encountering domain-specific event boundary tokens (e.g., comma `,' separators in BlocksWorld).
This design choice is grounded in the structure of sequential decision-making tasks:
\begin{itemize}[leftmargin=*]
    \item \textbf{Semantic Alignment: }Plan execution consists of variable-length sub-actions. Fixed-interval triggering risks updating the target latent prior $P_i$ mid-action, forcing the model to re-steer before a sub-goal is completed.
    \item \textbf{Variable-Length Robustness: }Event-driven triggers ensure that state updates naturally adapt to actions of varying token lengths without requiring artificial padding or manual segment alignment.
\end{itemize}

\section{Appendix B: dataset formalisation and generation algorithms}

To isolate sequence length, local search traps, and dynamic error recovery, our evaluation datasets were procedurally generated. Below, we formalise the generation logic and algorithms for the Treasure Hunt and Blocksworld domains.

\subsection{Procedural Treasure Hunt (semantic \& numeric memory)}
\label{sec:th_dataset}

The Treasure Hunt benchmark evaluates long-horizon retrieval of a semantic target ($t_{\text{item}}$) and a 4-digit numeric passkey ($c_{\text{pass}}$) across a distractor sequence of token length $H$. 

Let $\mathcal{I}$ be the set of $12$ distinct semantic items (e.g., \textit{Ruby Amulet}, \textit{Sapphire Ring}) and $C_{\text{pass}} \in [1000, 9999]$ be the numeric passkey space. Distractor sentences $s \in \mathcal{D}$ (e.g., \textit{``I walked down the long, dusty hallway.''}) are sampled without immediate consecutive repetition to construct a filler trajectory of length $H$. 

At sequence end, a fixed local trigger $w$ prompts the model to generate the terminal action $y$, requiring exact retrieval of both $c_{\text{pass}}$ and $t_{\text{item}}$.

\begin{algorithm}[htbp]
\caption{Treasure Hunt trajectory generation}
\label{alg:trajectory_generation}
\begin{algorithmic}[1]
\Require Target horizon $H$, Item set $\mathcal{I}$, Passcode bounds $[1000, 9999]$, Distractors $\mathcal{D}$
\Ensure Prompt $g$, Trajectory $X$, Target label $y$

\State $t_{\text{item}} \gets \textsc{RandomSample}(\mathcal{I})$
\State $c_{\text{pass}} \gets \textsc{RandomInteger}(1000, 9999)$
\State $g \gets \text{``[GOAL] Target: ''} \mathbin{\Vert} t_{\text{item}} \mathbin{\Vert} \text{`` | Code: ''} \mathbin{\Vert} c_{\text{pass}}$
\State $X \gets \emptyset, \quad \text{Available} \gets \textsc{Shuffle}(\mathcal{D})$

\While{$\textsc{TokenLength}(X) < H$}
    \If{$\text{Available is empty}$}
        \State $\text{Available} \gets \textsc{Shuffle}(\mathcal{D})$
    \EndIf
    \State $s \gets \textsc{Pop}(\text{Available})$
    \If{$X \neq \emptyset \textbf{ and } s = \textsc{Last}(X)$}
        \State $\textsc{PushToBottom}(\text{Available}, s)$
        \State $s \gets \textsc{Pop}(\text{Available})$
    \EndIf
    \State $X \gets X \mathbin{\Vert} s$
\EndWhile

\State $w \gets \text{``I checked my map and realized the vault was hidden in the Library.''}$
\State $y \gets \text{``I entered ''} \mathbin{\Vert} c_{\text{pass}} \mathbin{\Vert} \text{`` and retrieved the ''} \mathbin{\Vert} t_{\text{item}}$
\State \Return $g, \, X \mathbin{\Vert} w, \, y$
\end{algorithmic}
\end{algorithm}

\textbf{Trajectory Structure Example ($H=1024$):}
\begin{itemize}[leftmargin=*]
    \item \textbf{Goal Prompt ($g$):} \texttt{[GOAL] Target: Ruby Amulet | Code: 4892}
    \item \textbf{Distractor Trajectory ($X$):} \texttt{"I walked down the dusty hallway... [Repeated for H tokens]"}
    \item \textbf{Local Trigger ($w$):} \texttt{"I checked my map and realized the vault was hidden in the Library."}
    \item \textbf{Terminal Action ($y$):} \texttt{"I entered 4892 and retrieved the Ruby Amulet."}
\end{itemize}

\subsection{Deterministic planning and perturbations (Blocksworld)}
\label{sec:bw_dataset}

We model Blocksworld as a deterministic transition system $\Sigma = (\mathcal{S}, \mathcal{A}, \gamma)$, where $\mathcal{S}$ represents configurations of $N$ blocks, $\mathcal{A}$ is the action space (\texttt{pickup}, \texttt{putdown}, \texttt{stack}, \texttt{unstack}), and $\gamma: \mathcal{S} \times \mathcal{A} \to \mathcal{S}$ is the transition function.

Initial state $s_{\text{init}}$ and goal predicates $g$ are generated via random walks from a ground state (all blocks on table). Optimal paths $\tau_{\text{opt}}$ are solved via Breadth-First Search (BFS). 

To train and evaluate dynamic error recovery, Algorithm \ref{alg:trajectory_perturbation} programmatically injects up to $K$ legal sub-optimal actions into the execution sequence, recalculating optimal recovery paths on the fly.

\begin{algorithm}[h]
\caption{Blocksworld $K$-perturbation trajectory generation}
\label{alg:trajectory_perturbation}
\begin{algorithmic}[1]
\Require Initial state $s_{\text{init}}$, Goal conditions $g$, Max perturbations $K$
\Ensure Formatted trajectory $\tau$, Boolean flag $\textit{is\_perturbed}$

\State $\tau_{\text{opt}} \gets \textsc{BFS}(s_{\text{init}}, g)$
\If{$\tau_{\text{opt}} = \emptyset \textbf{ or } K = 0$}
    \State \Return $\tau_{\text{opt}}, \, \text{False}$
\EndIf

\State $s_{\text{curr}} \gets s_{\text{init}}, \quad \tau \gets \emptyset, \quad m \gets 0$

\While{$s_{\text{curr}} \not\models g$}
    \State $\tau_{\text{rem}} \gets \textsc{BFS}(s_{\text{curr}}, g)$
    \State $a_{\text{opt}} \gets \textsc{First}(\tau_{\text{rem}})$
    
    \If{$m < K \textbf{ and } \textsc{Random}(0, 1) < 0.3$} \Comment{Inject perturbation}
        \State $\mathcal{V}_{\text{valid}} \gets \textsc{GetValidActions}(s_{\text{curr}})$
        \State $\mathcal{V}_{\text{err}} \gets \mathcal{V}_{\text{valid}} \setminus \{a_{\text{opt}}\}$
        \If{$\mathcal{V}_{\text{err}} \neq \emptyset$}
            \State $a \gets \textsc{RandomSample}(\mathcal{V}_{\text{err}})$
            \State $m \gets m + 1$
        \Else
            \State $a \gets a_{\text{opt}}$
        \EndIf
    \Else
        \State $a \gets a_{\text{opt}}$
    \EndIf
    
    \State $\tau \gets \tau \mathbin{\Vert} \{a\}$
    \State $s_{\text{curr}} \gets \gamma(s_{\text{curr}}, a)$ \Comment{Transition to new state}
\EndWhile

\State \Return $\tau, \, (m > 0)$
\end{algorithmic}
\end{algorithm}

\subsection{Implementation and reproducibility details}
\label{sec:bw_implementation}
\subsubsection{Domain-specific tokenisation}
\label{sec:bw_tokenization}
A known confounding factor when evaluating logical reasoning in language models is Byte-Pair Encoding (BPE) fragmentation \citep{chai2024tokenization, zhang2024counting}. Standard tokenisers frequently split single-character identifiers (e.g., blocks A through F) or fuse structural delimiters with adjacent action verbs, artificially complicating state tracking. To isolate architectural capacity from lexical artifacts, we adopt a clean word-level tokenisation scheme for the Blocksworld domain.

We map environmental entities, action primitives (\texttt{pickup}, \texttt{putdown}, \texttt{stack}, \texttt{unstack}), and relational predicates directly to individual vocabulary IDs. Key structural delimiters, including \texttt{[INIT]}, \texttt{[GOAL]}, \texttt{[DONE]}, and the update trigger comma (\texttt{,}), are registered as immutable special tokens. This 1:1 mapping ensures that the latent steering vector $P_i$ interacts with clean, unfragmented physical semantics.

\subsubsection{VICReg formulation and hyperparameters}
\label{sec:vicreg_hyper}
The Latent Predictor $E_{\text{pred}}$ is implemented as a 3-layer Multi-Layer Perceptron (MLP) with GELU activations matching the base model hidden dimension ($d=512$). During Phase 2 alignment, $E_{\text{pred}}$ is trained using the Variance-Invariance-Covariance Regularisation (VICReg) objective \citep{bardes2021vicreg} over batch size $B=32$ across representation matrices $P_0, Z_{\text{future}} \in \mathbb{R}^{B \times d}$:
$$ \mathcal{L}_{\text{align}} = \lambda_s s(P_0, Z_{\text{future}}) + \lambda_v \Big[ v(P_0) + v(Z_{\text{future}}) \Big] + \lambda_c \Big[ c(P_0) + c(Z_{\text{future}}) \Big] $$
where the individual objective terms are defined as:
\begin{itemize}[leftmargin=*]
    \item \textbf{Invariance ($s$):} Mean squared error between batch predictions and target representations: $s(P_0, Z_{\text{future}}) = \frac{1}{B} \sum_{i=1}^B \| P_{0,i} - Z_{\text{future},i} \|_2^2$.
    \item \textbf{Variance ($v$):} Hinge loss maintaining the standard deviation of each feature dimension above target margin $\gamma = 1.0$: $v(Z) = \frac{1}{d} \sum_{j=1}^d \max\big(0, \, \gamma - \text{std}_j(Z)\big)$.
    \item \textbf{Covariance ($c$):} Sum of squared off-diagonal terms of the feature covariance matrix to decorrelate latent channels: $c(Z) = \frac{1}{d} \sum_{i \neq j} [C(Z)]_{i,j}^2$.
\end{itemize}
We adopt standard VICReg weighting parameters: $\lambda_s = 25.0$, $\lambda_v = 25.0$, and $\lambda_c = 1.0$, with standard deviation numerical stability constant $\epsilon = 10^{-4}$.

\subsubsection{Training hyperparameters: }
\label{sec:train_hyper}
\paragraph{Hardware and computational budget}All models were implemented in PyTorch. Experiments were conducted on an internal cluster using NVIDIA A100 (80GB VRAM) GPUs (utilising PyTorch Distributed Data Parallel for multi-GPU runs).
Individual training runs required approximately 3--5 wall-clock hours for 32M parameter models and 5--8 wall-clock hours for 85M parameter models. Accounting for preliminary hyperparameter tuning, loss stability testing, and exploratory/failed iterations, the total computational budget for the entire project is estimated at approximately 350 A100 GPU hours.

Models were trained using the AdamW optimizer ($\beta_1 = 0.9, \beta_2 = 0.999$, weight decay $= 1 \times 10^{-4}$) with a global gradient clipping norm of 1.0. The detailed training schedules across experimental runs are as follows:
\begin{itemize}[leftmargin=*]
    \item \textbf{Phase 1 (Base Domain Adaptation):} The base model ($d=512$, $L=4$ layers) is fine-tuned for 30 epochs at a peak learning rate of $1 \times 10^{-4}$ with a 10\% linear warmup and linear decay.
    \item \textbf{Phase 2 (Predictor Alignment):} $E_{\text{pred}}$ is trained for 30 epochs at a peak learning rate of $2 \times 10^{-4}$ with batch size $B=32$, keeping base encoder $E_{\text{base}}$ frozen.
    \item \textbf{Phase 3 (Conditioned Generation):} The causal decoder $D_{\text{base}}$, zero-initialised FiLM projections ($W_\gamma, W_\beta, b_\gamma, b_\beta$), and the Dynamic State Tracker are trained end-to-end for 30 epochs at a uniform learning rate of $1 \times 10^{-4}$. Textual goal dropout is applied with probability $p_{\text{drop}} = 0.30$. The Tracker gate scalar is initialised to $\alpha = 0.0$.
\end{itemize}

All sequence generation during evaluation uses deterministic greedy decoding ($\tau = 0.0$).


\end{document}